\documentclass[11pt]{article}

\usepackage{acl}
\usepackage{times}
\usepackage{latexsym}

\usepackage[protrusion=false]{microtype}

\usepackage{graphicx}
\usepackage{booktabs}
\usepackage{amsmath}
\usepackage{tikz}
\usepackage{pgfplots}

\usepgfplotslibrary{groupplots}
\pgfplotsset{compat=1.18}
\usetikzlibrary{arrows.meta, positioning}

\title{Investigating Temporal Motion Features for Pose-to-Text Indian Sign Language Translation}

\author{Manav Dhamecha, Praveen Kumar Chandaliya, Pruthwik Mishra \\
  Sardar Vallabhbhai National Institute of Technology, Surat, India\\
  \texttt{\{u24ai034, pkc, pruthwikmishra\}@aid.svnit.ac.in}
  }

\begin{document}

\nolinenumbers
\pagestyle{empty}

\maketitle

\begin{abstract}
We investigate the effect of pretrained T5 model scale and explicit motion features on pose-to-text Indian Sign Language Translation (SLT) for the WSLP 2026 Shared Task. Pose sequences are projected into the embedding space of T5 through a lightweight pose encoder, with the complete model fine-tuned to generate English text. The shared task data used for this work consists of a test set with 5,334 examples and a validation set with 5,257 examples. We compare T5-small, T5-base, and T5-large, and additionally introduce a motion-augmented variant, T5-small + Motion, that adds explicit frame-to-frame pose differences to the input representation. T5-small achieves the best BLEU and ROUGE scores among the spatial-only models, while T5-large obtains the highest chrF score. Augmenting T5-small with motion features yields the largest single improvement observed in our study, substantially improving BLEU over the spatial-only baseline and making it the strongest model overall on this metric. Our submitted system ranked 5th on the official WSLP 2026 SLT testing leaderboard. The source code and trained models are publicly available on
GitHub~\footnote{\url{https://github.com/manavdhamecha77/Sign-Lang-Trans}}
and HuggingFace~\footnote{\url{https://huggingface.co/manavdhamecha77/iSign-t5-pose-to-text}}
respectively. Our findings provide an empirical analysis of pose-conditioned T5 models for Indian Sign Language translation in a shared-task setting.
\end{abstract}

\section{Introduction}

Sign language translation (SLT) aims to map visual sign language sequences
into spoken or written language \citep{camgoz2018neural,camgoz2020sign}.
Recent work has explored transformer-based approaches and the reuse of
pretrained language models for low-resource translation settings
\citep{yin-read-2020-better,decoster2021frozen,zhang-duh-2021-approaching}.
For Indian Sign Language, the iSign corpus provides a large resource for
studying recognition and translation \citep{joshi-etal-2023-isign}.

In this work, we use the iSign poses v1.1 Part AD subset, containing pose
representations for 18,867 sign language samples. The subset consists of 16,980 samples used for model training and
1,887 samples used as a held-out evaluation split in our experiments. Each sample is represented as a sequence of body and hand keypoints, which serves as the input to our
pose-based sign language translation models.

Instead of processing raw video, pose-based SLT represents each frame through
body and hand keypoints. This removes visual background variation and provides
a compact structured representation suitable for sequence models
\citep{moryossef2021evaluating,zelezny2025exploring}. Recent work
\cite{zelezny2025exploring} further investigates pose-based SLT through
ablation studies and attention analysis using a T5-style encoder--decoder
framework. Building on this direction, we investigate two complementary
questions: (i) whether larger pretrained text-to-text transformers provide a
consistent advantage when adapted to this non-textual input modality, and
(ii) whether enriching the spatial pose representation with explicit motion
information is a more effective use of model capacity than simply scaling
the language model.

We project pose sequences into the input embedding space of pretrained T5
models \citep{raffel2020exploring} and fine-tune three scale variants,
T5-small, T5-base, and T5-large, as well as a motion-augmented variant,
T5-small + Motion, which concatenates frame-to-frame pose differences with
the spatial pose features. All four models are compared on the same held-out
10\% split of the iSign corpus. Separately, our best submission was scored
on the official WSLP 2026 SLT \texttt{test.csv} split and ranked 5th on the
official testing leaderboard.

Our main findings are:

\begin{itemize}
\item T5-small achieves the best BLEU, ROUGE-1, ROUGE-2, and ROUGE-L
scores among the spatial-only scale variants.

\item T5-large achieves the best chrF score but receives only three
training epochs compared with ten for T5-small and T5-base.

\item Model scaling from T5-small to T5-large does not produce a
monotonic improvement under our experimental setup, highlighting the
importance of matched optimization and training budgets.

\item Augmenting spatial pose representations with explicit motion
features substantially improves translation performance, with
T5-small + Motion achieving the best BLEU score of 0.298, compared with
0.189 for the spatial-pose T5-small baseline---a relative improvement of approximately 58\% in BLEU, while chrF also improves by approximately 2.0\%.
\end{itemize}

\section{Related Work}

Neural SLT has evolved from recurrent sequence-to-sequence systems toward
transformer-based architectures \citep{camgoz2018neural,camgoz2020sign}.
Transformer models have also been successfully applied to sign or gloss-to-text
translation \citep{yin-read-2020-better}. Another direction reuses pretrained
language models by projecting non-textual or visual features into a language
model representation space \citep{decoster2021frozen}.

Pose estimation offers an alternative to raw video by representing sign motion
through structured landmarks. \citet{moryossef2021evaluating} studied the use
of pose estimation for sign language processing and introduced tools for
working with pose representations. More recently, \citet{zelezny2025exploring}
presented a pose-based SLT framework with ablation studies and attention
analysis using a T5-style encoder--decoder model. Our implementation is based on the conversion of pose-to-text while adapting it to Indian Sign Language
using iSign \citep{joshi-etal-2023-isign} and studies the
effect of T5 model scale and explicit motion features derived from the pose
sequence.

\section{Methodology}

\subsection{Pose Representation}

We process the provided sign-language pose sequences using the Python
\texttt{pose-format} library\footnote{\url{https://pypi.org/project/pose-format/}}. The \texttt{.pose} files are loaded and
converted into frame-wise landmark representations containing body,
left-hand, and right-hand keypoints. We use 3D coordinates and confidence
values for each landmark, resulting in a 300-dimensional spatial feature
vector per frame.

The representation contains 33 body landmarks and 21 landmarks for each
hand. Each landmark provides three spatial coordinates and a confidence
value. Sequences are truncated or padded to 500 frames and normalized per
example.

\subsection{Motion Features}
\label{sec:mot_feat}
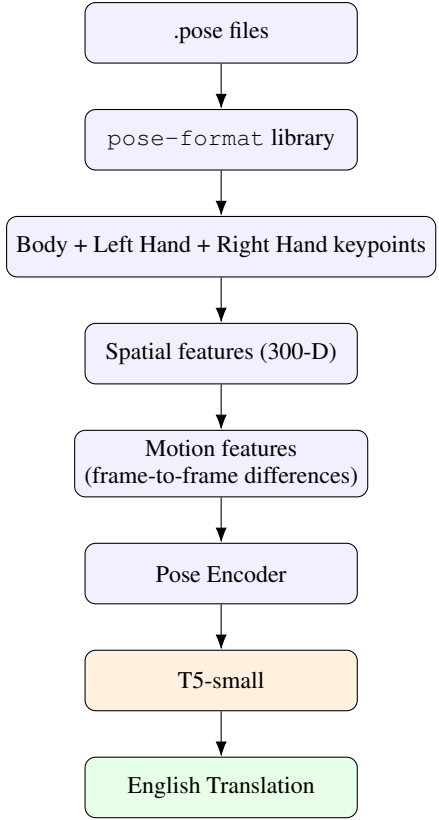
\begin{figure}[t]
\centering
\begin{tikzpicture}[
  node distance=6mm,
  every node/.style={font=\small},
  block/.style={rectangle, draw, rounded corners, align=center,
                minimum width=3.6cm, minimum height=8mm, fill=blue!6},
  arr/.style={-{Latex[length=2mm]}}
]
\node[block] (pose) {.pose files};
\node[block, below=of pose] (pf) {\texttt{pose-format} library};
\node[block, below=of pf] (kp) {Body + Left Hand + Right Hand keypoints};
\node[block, below=of kp] (spatial) {Spatial features (300-D)};
\node[block, below=of spatial] (motion) {Motion features\\(frame-to-frame differences)};
\node[block, below=of motion] (enc) {Pose Encoder};
\node[block, below=of enc, fill=orange!12] (t5) {T5-small};
\node[block, below=of t5, fill=green!10] (out) {English Translation};

\draw[arr] (pose) -- (pf);
\draw[arr] (pf) -- (kp);
\draw[arr] (kp) -- (spatial);
\draw[arr] (spatial) -- (motion);
\draw[arr] (motion) -- (enc);
\draw[arr] (enc) -- (t5);
\draw[arr] (t5) -- (out);
\end{tikzpicture}
\caption{End-to-end pipeline from raw \texttt{.pose} files to English text.
Spatial and motion features are concatenated before being passed to the
pose encoder.}
\label{fig:pipeline}
\end{figure}

Static per-frame landmarks capture the spatial configuration of a sign but do
not explicitly encode how that configuration changes over time. This is a
central component of sign articulation. We therefore compute an explicit
motion feature for each frame as the first-order temporal difference between
consecutive spatial feature vectors,
\begin{equation}
    m_t = x_t - x_{t-1},
\end{equation}
where $x_t$ is the 300-dimensional spatial feature vector at frame $t$ and
$m_0$ is set to zero for the first frame of a sequence. The motion feature
$m_t$ is concatenated with the spatial feature $x_t$, giving a 600-dimensional
per-frame input to the pose encoder. This provides the model with explicit
frame-wise velocity information besides the absolute pose, without
requiring any additional annotation or a change to the underlying pose
information. Figure~\ref{fig:pipeline} summarizes the resulting end-to-end pipeline.

\subsection{Pose-Conditioned T5}

A lightweight multi-layer perceptron pose encoder maps each per-frame feature vector (300 dimensions for
the spatial-only models and 600 dimensions for the motion-augmented model) to the hidden
dimension of the corresponding T5 model. The encoder consists of a linear
layer, a 256-dimensional hidden representation, ReLU activation, dropout, and
a projection to $d_{\text{model}}$.

The resulting sequence is passed to T5 through its input embeddings.
An attention mask prevents padded frames from contributing to self-attention.
The pose encoder and all T5 parameters are fine-tuned jointly using
teacher-forced text generation.

\section{Experimental Setup}

We use the WSLP 2026 SLT shared task data, with pose sequences read directly
from the provided \texttt{.pose} files using the Python \texttt{pose-format}
library. This provides a structured representation of the body and hand
keypoints used as input to the pose encoder. The official test and validation splits for the shared task contain 5,334 examples and 5,257 examples, respectively, in \texttt{.pose} files. The remaining released data is
used for training. The final
leaderboard ranking is based on the scores obtained on the test set.

For the model comparison in this paper (Table~\ref{tab:results}), we instead
follow common practice for controlled ablation and hold out a fixed 10\%
split of the iSign corpus \citep{joshi-etal-2023-isign} (1,887 examples) from
our own training data. All four models -- T5-small, T5-base, T5-large, and
T5-small + Motion -- are trained and compared on this same iSign 10\%
held-out split throughout the paper, which is distinct from the official
WSLP 2026 test set. We use this internal split so that all four
models can be compared under identical conditions with unlimited evaluation
budget, independent of the shared task's leaderboard scoring schedule. This held-out validation set is not used in training the models.

We evaluate T5-small ($\sim$60M parameters), T5-base ($\sim$220M), and
T5-large ($\sim$770M) \citep{raffel2020exploring} on spatial pose features
only, and a fourth model, T5-small + Motion, which uses the concatenated
spatial and motion representation described in Section~\ref{sec:mot_feat}.

All models use AdamW \cite{loshchilov2018decoupled}, batch size 16, gradient accumulation of 2, gradient
clipping at 1.0, fp16 training, and a learning rate of $5\times10^{-4}$.
T5-small, T5-base, and T5-small + Motion are trained for 10 epochs. T5-large
is trained for 3 epochs because of its substantially higher computational
cost.

For inference, we use beam search with beam size 4, length penalty 2.0, and
a maximum generation length of 128 tokens. We report BLEU
\citep{papineni2002bleu}, chrF \citep{popovic-2015-chrf}, and ROUGE
\citep{lin-2004-rouge} on the same iSign 10\% held-out split.

\section{Results and Discussion}

\begin{figure*}[t]
\centering

\begin{tikzpicture}

\begin{groupplot}[
    group style={
        group size=2 by 1,
        horizontal sep=2cm
    },
    width=7cm,
    height=5cm,
    ybar,
    symbolic x coords={
        T5-small,
        T5-base,
        T5-large,
        T5-small+Motion
    },
    xtick=data,
    x tick label style={
        rotate=35,
        anchor=east,
        font=\small
    },
    nodes near coords,
    every node near coord/.append style={
        font=\small
    },
    enlarge x limits=0.2
]


\nextgroupplot[
    title={BLEU},
    ylabel={BLEU},
    ymin=0,
    ymax=0.35,
    bar width=12pt
]

\addplot coordinates {
    (T5-small,0.189)
    (T5-base,0.154)
    (T5-large,0.132)
    (T5-small+Motion,0.298)
};


\nextgroupplot[
    title={chrF},
    ylabel={chrF},
    ymin=0,
    ymax=18,
    bar width=12pt
]

\addplot coordinates {
    (T5-small,14.44)
    (T5-base,12.81)
    (T5-large,15.59)
    (T5-small+Motion,14.73)
};

\end{groupplot}

\end{tikzpicture}

\caption{
BLEU (left) and chrF (right) across the four evaluated models on the
1,887-example iSign 10\% held-out split. T5-small + Motion obtains the
highest BLEU, while T5-large achieves the highest chrF.
}

\label{fig:barchart}

\end{figure*}
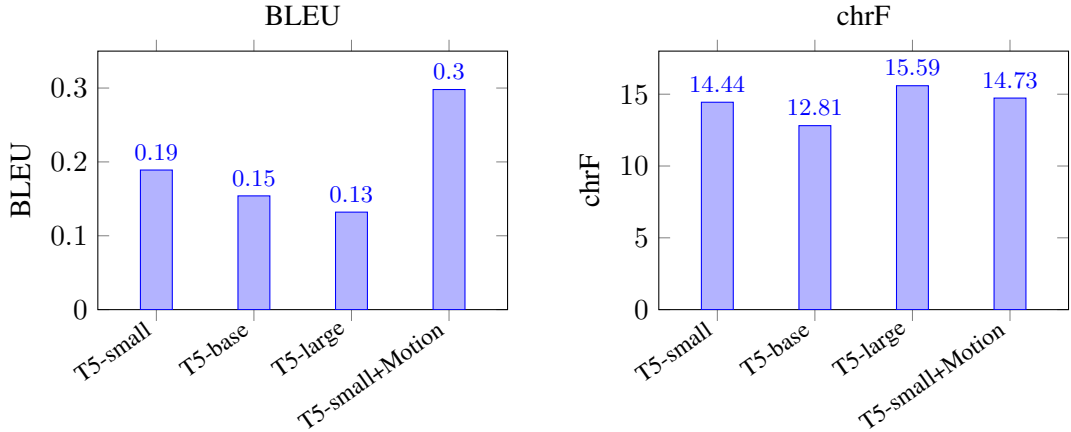

Table~\ref{tab:results} presents the final results on all 1,887 examples of
the iSign 10\% held-out split, and Figure~\ref{fig:barchart} visualizes BLEU
and chrF across the four models on this same split.

\begin{table}[t]
\centering
\small
\resizebox{\columnwidth}{!}{%
\begin{tabular}{lrrrrr}
\toprule
Model & BLEU & chrF & R-1 & R-2 & R-L \\
\midrule
T5-small & 0.189 & 14.44 & \textbf{9.13} & 0.60 & \textbf{8.14} \\
T5-base  & 0.154 & 12.81 & 6.20 & 0.51 & 5.51 \\
T5-large & 0.132 & \textbf{15.59} & 8.71 & 0.25 & 8.02 \\
T5-small + Motion & \textbf{0.298} & 14.73 & 8.69 & \textbf{0.77} & 7.67 \\
\bottomrule
\end{tabular}%
}
\caption{Results on the 1,887-example iSign 10\% held-out split. ROUGE
values are shown as percentages for readability. Bold indicates the best score per
column.}
\label{tab:results}
\end{table}

T5-small performs best on four of the five main evaluation measures among
the spatial-only models. It obtains the highest BLEU score of 0.189 among
these three and the strongest ROUGE-1 and ROUGE-L scores, indicating better
word and sequence overlap with the reference translations. T5-large achieves
the highest chrF score of 15.59, suggesting relatively better
character-level overlap, but its BLEU and ROUGE scores remain below those of
T5-small.

Among the spatial-only variants, the results do not show a monotonic benefit
from increasing model size. T5-base performs worst on all reported metrics
except ROUGE-2, while T5-large does not consistently outperform the smaller
models despite its larger capacity. However, these results should not be
interpreted as proving that smaller models are inherently better for
pose-to-text SLT. T5-large was trained for only three epochs, whereas
T5-small and T5-base were trained for ten. Additionally, the training
configuration used a constant learning rate of $5\times10^{-4}$ without the
intended warmup schedule being applied. These factors may disproportionately
affect larger pretrained models.

Adding explicit motion features to T5-small produces the largest single
improvement observed in this study. BLEU rises from 0.189 to 0.298,
corresponding to a relative improvement of approximately 58\% over the
spatial-only T5-small baseline. The chrF score also increases from 14.44 to
14.73, corresponding to a relative improvement of approximately 2.0\%.
ROUGE-2 also improves from 0.60 to 0.77. However, T5-large achieves the
highest overall chrF score. ROUGE-1 and ROUGE-L decrease slightly relative
to the spatial-only T5-small model, suggesting that motion information
primarily benefits local n-gram fidelity captured by BLEU rather than the
longest-common-subsequence style overlap measured by ROUGE-L. These results
suggest that, under the evaluated configuration, improving the temporal
richness of the pose representation is a more effective use of the fixed
T5-small capacity than increasing the number of parameters in the underlying
language model.

The low absolute scores across all models also indicate that the main
challenge lies beyond model scaling. The pose encoder is trained from
scratch, and learning a direct mapping from continuous pose sequences to a
pretrained language model embedding space is indeed challenging. 

\section{Conclusion}

We present a comparison of T5-small, T5-base, and T5-large for
pose-to-text Indian Sign Language Translation using pose sequences, together
with a motion-augmented T5-small variant. Under the evaluated configuration,
T5-small achieves the best BLEU and ROUGE scores among the spatial-only
models, while T5-large obtains the best chrF score. Model scaling from
T5-small to T5-large does not consistently improve translation performance.
In contrast, augmenting T5-small with explicit frame-to-frame motion features
substantially improves BLEU. This suggests that, in our evaluated setup,
improving the temporal richness of the pose representation may be more
effective than simply increasing the size of the underlying T5 model.

Because T5-large received fewer training epochs and the intended
learning-rate warmup was not applied, the scale comparison should be viewed
as a comparison of the evaluated training configurations rather than a
definitive conclusion about model capacity. Future work should apply
motion-augmented representations to larger T5 variants under matched training
budgets, use improved optimization schedules, and explore additional temporal
or velocity-based pose features.

\section{Limitations}

Our experiments use a single dataset and a single training run per model.
The four models do not receive identical training budgets -- T5-large in
particular is trained for fewer epochs -- and we do not report multi-seed
variance or human evaluation. The motion feature is a simple first-order
difference and has only been evaluated in combination with T5-small; we do
not yet know whether the same improvement transfers to other variants such as T5-base or T5-large.
Furthermore, all models achieve low absolute automatic metric scores, so the
resulting systems should be considered research baselines rather than
deployable translation systems.

\section*{Ethics Statement}

This work uses pose data provided for the WSLP 2026 SLT shared task,
processed with the openly available \texttt{pose-format} library, and builds
on the publicly available iSign corpus \citep{joshi-etal-2023-isign}. We
study pose representations rather than raw video and do not collect new
participant data. Because incorrect translations can lead to
misunderstanding, the models presented here should not be used in
accessibility-critical settings without substantial further validation.
\section*{Acknowledgment}
The authors acknowledge the Department of Computer Science and Engineering and the Department of Artificial Intelligence, SVNIT Surat, for providing access to the Dual NVIDIA H100 NVL GPU Computing Cluster used in this research.
\bibliography{references}

\end{document}